\documentclass{article}
\usepackage[preprint]{spconf}

\title{
Language Modeling for Multi-Domain Speech-Driven Text Retrieval
}

\name{
Katunobu Itou$^1$, Atsushi Fujii$^2$, Tetsuya Ishikawa$^2$
\thanks{The first and second authors
       are also members of CREST, Japan Science and Technology
       Corporation.}
}

\address{
$^1$ National Institute of Advanced Industrial Science and Technology\\
  1-1-1 Chuuou Daini Umezono, Tsukuba, 305-8568, Japan, 
E-mail: itou@ni.aist.go.jp\\
$^2$ University of Library and Information Science\\
      1-2 Kasuga, Tsukuba, 305-8550, Japan, 
      E-mail: \{fujii,ishikawa\}@ulis.ac.jp \\
}

\begin{document}
\ninept
\maketitle
\begin{abstract}
We report experimental results associated with speech-driven text
retrieval, which facilitates retrieving information in multiple
domains with spoken queries. Since users speak contents related to a
target collection, we produce language models used for speech
recognition based on the target collection, so as to improve both the
recognition and retrieval accuracy. Experiments using existing test
collections combined with dictated queries showed the effectiveness of
our method.
\end{abstract}

\newcommand{\etal}{et~al.}
\newcommand{\etaleos}{et~al}
\newcommand{\eq}[1]{(\ref{#1})}
\input{psfig.tex}

\section{Introduction}
\label{sec:introduction}

Automatic speech recognition, which decodes human voice to generate
transcriptions, has of late become a practical technology.  It is
feasible that speech recognition is used in real world computer-based
applications, specifically, those associated with human language.  In
fact, a number of speech-based methods have been explored in the
information retrieval (IR) community, which can be classified into the
following two fundamental categories:
\begin{itemize}
\item spoken document retrieval, in which written queries are used to
  search speech (e.g., broadcast news audio) archives for relevant
  speech information~\cite{garofolo:trec-97}.
\item speech-driven retrieval, in which spoken queries are used to
  retrieve relevant textual information~\cite{barnett:eurospeech-97,crestani:fqas-2000}.
\end{itemize}

Initiated partially by the TREC-6 spoken document retrieval (SDR)
track~\cite{garofolo:trec-97}, various methods have been proposed for
spoken document retrieval.  However, a relatively small number of
methods have been explored for speech-driven text retrieval, although
they are associated with numerous keyboard-less retrieval
applications, such as telephone-based retrieval, car navigation
systems, and user-friendly interfaces.

Barnett~\etal~\cite{barnett:eurospeech-97} performed comparative
experiments related to speech-driven retrieval, where the DRAGON
speech recognition system was used as an input interface for the
INQUERY text retrieval system.  They used as test inputs 35 queries
collected from the TREC topics and dictated by a single male speaker.
Crestani~\cite{crestani:fqas-2000} also used the above 35 queries and
showed that conventional relevance feedback techniques marginally
improved the accuracy for speech-driven text retrieval.

These above cases focused solely on improving text retrieval methods
and did not address problems of improving speech recognition accuracy.
In fact, an existing speech recognition system was used with no
enhancement. In other words, speech recognition and text retrieval
modules were fundamentally independent and were simply connected by
way of an input/output protocol.

However, since most speech recognition systems are trained based on
specific domains, the accuracy of speech recognition across domains is
not satisfactory. Thus, as can easily be predicted, in cases of
Barnett~\etal~\cite{barnett:eurospeech-97} and
Crestani~\cite{crestani:fqas-2000}, a speech recognition error rate
was relatively high and considerably decreased the retrieval accuracy.
Additionally, speech recognition with a high accuracy is crucial for
interactive retrieval, such as dialog-based retrieval.

Motivated by these problems, in this paper we integrate (not simply
connect) speech recognition and text retrieval to improve both
recognition and retrieval accuracy in the context of speech-driven
text retrieval.

Unlike general-purpose speech recognition aimed to decode any
spontaneous speech, in the case of speech-driven text retrieval, users
usually speak contents associated with a target collection, from which
documents relevant to their information need are retrieved.  In a
stochastic speech recognition framework, the accuracy depends
primarily on acoustic and language models~\cite{bahl:ieee-tpami-1983}.
While acoustic models are related to phonetic properties, language
models, which represent linguistic contents to be spoken, are
related to target collections.  Thus, it is intuitively feasible that
language models have to be produced based on target collections.

To sum up, our belief is that by adapting a language model based on a
target IR collection, we can improve the speech recognition and text
retrieval accuracy, simultaneously.

Section~\ref{sec:system} describes our speech-driven text retrieval
system, which is currently implemented for Japanese.
Section~\ref{sec:experimentation} elaborates on comparative
experiments, in which IR test collections in different domains are
used to evaluate the effectiveness of our system.

\section{System Description}
\label{sec:system}

\subsection{Overview}
\label{subsec:system_overview}

\begin{figure}[htbp]
  \begin{center}
  \leavevmode \psfig{file=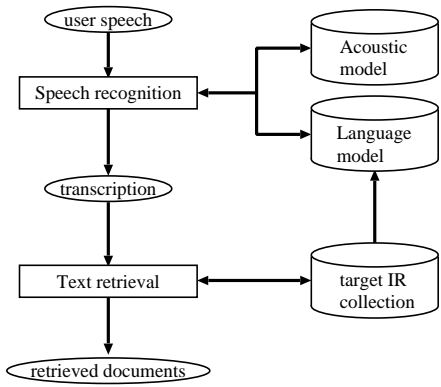,height=2in}
  \end{center}
  \caption{The design of our speech-driven text retrieval system.}
  \label{fig:system}
\end{figure}

Figure~\ref{fig:system} depicts the overall design of our
speech-driven text retrieval system, which consists of speech
recognition and text retrieval modules. 
In the following sections, we explain two modules in
Figure~\ref{fig:system}, respectively.

\subsection{Speech Recognition}
\label{subsec:speech_recognition}

For the speech recognition module, we use the Japanese dictation
toolkit~\cite{kawahara:icslp-2000}\footnote{http://winnie.kuis.kyoto-u.ac.jp/dictation/},
which includes the ``Julius'' recognition engine and acoustic/language
models.  Julius performs a two-pass (forward-backward) search using
word-based forward bigrams and backward trigrams on the respective
passes.

The acoustic model was produced by way of the ASJ speech databases of
phonetically balanced sentences (ASJ-PB) and newspaper articles texts
(ASJ-JNAS)~\cite{itou:98:a}, which contain approximately 20,000
sentences uttered by 132 speakers including the both gender groups.
We used a 16-mixture Gaussian distribution triphone
Hidden Markov Model, where states were clustered into 2,000 groups by
a state-tying method.

This toolkit also includes development softwares, so that acoustic and
language models can be produced and replaced depending on the
application.  While we use the acoustic model provided in the toolkit,
we use new language models produced by way of source documents (i.e.,
target IR collections).

\subsection{Text Retrieval}
\label{subsec:text_retrieval}

The text retrieval module is based on the ``Okapi''
method~\cite{robertson:sigir-94}, which computes the relevance score
between the transcribed query and each document in the collection,
based on the distribution of index terms, and sorts retrieved documents
according to the score in descending order.

We use content words extracted from documents as index terms, and
perform a word-based indexing. For this purpose, we use the ChaSen
morphological analyzer~\cite{matsumoto:chasen-99} to extract content
words. We extract terms from transcribed queries using the same
method.

\section{Experimentation}
\label{sec:experimentation}

\subsection{Test Collections}
\label{subsec:test_collection}

To investigate the performance of our multi-domain speech-driven
retrieval system, we used two different types of Japanese IR test
(benchmark) collections: the NTCIR and IREX collections. Both
collections, which resemble one used in the TREC ad hoc retrieval
track, include topics (information need) and relevance assessment
(correct judgement) for each topic, along with target
documents. However, these collections are associated with different
domain, respectively.

The NTCIR
collection~\cite{ntcir-2001}\footnote{http://research.nii.ac.jp/\~{}ntcadm/index-en.html}
includes 736,166 abstracts collected from technical papers published
by 65 Japanese associations for various fields.  On the other hand,
the IREX
collection~\cite{sekine:lrec-2000}\footnote{http://cs.nyu.edu/cs/projects/proteus/irex/index-e.html}
includes 211,853 articles collected from two years worth of ``Mainichi
Shimbun'' newspaper articles\footnote{In practice, the IREX collection
provides only article IDs, which corresponds to articles in Mainichi
Shimbun newspaper CD-ROM'94-'95. Participants must get a copy of the
CD-ROMs themselves.}.

The NTCIR and IREX collections include 132 and 30 Japanese topics,
respectively, for a sample of which English translations are also
provided. Figures~\ref{fig:ntcir_topic} and \ref{fig:irex_topic} show
example topics in each collection, which consist of different fields
(for example, descriptions and narratives) tagged in an SGML form.

\begin{figure*}[htbp]
  \begin{center}
    \leavevmode
    \begin{quote}
      \tt
      \footnotesize
      <TOPIC q=0123>\\
      <TITLE>Biofilms</TITLE>\\
      <DESCRIPTION>Are there any documents about the biofilms produced
      by some microorganisms in which chronic diseases are mentioned?</DESCRIPTION>\\
      <NARRATIVE>Biofilms are thought to occur when microorganisms
      grow in microcolonies embedded in the adherent gel surface on
      tunica mucosa, and teeth, or on catheters, prosthetic valves,
      and other artifacts. A relevant document will report on any
      studies into the relationship between biofilms produced by some
      microorganisms and chronic diseases. Documents that include
      reports on biofilms produced by non-medical microorganisms that
      do not cause infectious diseases are not relevant.</NARRATIVE>\\
      </TOPIC>
    \end{quote}
    \caption{An English translation for an example topic in the NTCIR
      collection.}
    \label{fig:ntcir_topic}
  \end{center}
\end{figure*}

\begin{figure*}[htbp]
  \begin{center}
    \leavevmode
    \begin{quote}
      \tt
      \footnotesize
      <TOPIC> \\
      <TOPIC-ID>1001</TOPIC-ID> \\
      <DESCRIPTION>Corporate merging</DESCRIPTION> \\
      <NARRATIVE>The article describes a corporate merging and in the
      article, the name of companies have to be
      identifiable. Information
      including the field and the purpose of the merging have to be
      identifiable. Corporate merging includes corporate acquisition,
      corporate unifications and corporate buying.</NARRATIVE> \\
      </TOPIC>
    \end{quote}
    \caption{An English translation for an example topic in the IREX collection.}
    \label{fig:irex_topic}
  \end{center}
\end{figure*}

Since both collections do not contain spoken queries, we asked four
speakers (two males/females) to dictate topics. For this purpose, we
selectively used a specific field, so as to simulate a realistic
speech-driven retrieval.

In the case of the NTCIR topics, titles are not informative for the
retrieval. On the other hand, narratives, which usually consist of
several sentences, are too long to speak. Thus, only descriptions,
which consist of a single phrase and sentence, were dictated by each
speaker, so as to produce four different sets of 132 spoken queries.
However, in the case of the IREX topics, since descriptions are not
informative for the retrieval, only narratives were dictated by each
speaker, to produce four different sets of 30 spoken queries.

\subsection{Comparative Evaluation}
\label{subsec:comparison}

We compared the performance of the following retrieval methods:
\begin{itemize}
\item text-to-text retrieval, which used written queries, and can be
  seen as the perfect speech-driven text retrieval,
\item speech-driven text retrieval, in which a language model produced
  based on the NTCIR collection was used,
\item speech-driven text retrieval, in which a language model produced
  based on the IREX collection was used.
\end{itemize}
In cases of speech-driven text retrieval methods, queries dictated by
four speakers were used independently, and the final result was
obtained by averaging results for different speakers.

Although the Julius decoder outputs more than one transcription
candidates for a single speech, we used only the one with the greatest
probability score. The results did not significantly change depending
on whether or not we used lower-ranked transcriptions as queries.

The only difference in producing two different language models (i.e.,
those based on the NTCIR and IREX collections) is the source
documents. In other words, both language models were of the same
vocabulary size (20,000), and were produced by way of the same
softwares.

Table~\ref{tab:lang_model} shows statistics related to word
tokens/types in two different collections for language modeling, where
the line ``Coverage'' denotes the ratio of word tokens contained in
the resultant language model. Most of word tokens were covered
irrespective of the collection.

\begin{table}[htbp]
  \begin{center}
    \caption{Statistics related to source words for language
    modeling.}
    \medskip
    \leavevmode
    \tabcolsep=3pt
    \begin{tabular}{lcc} \hline
      & NTCIR & IREX \\ \hline
      \# of Types & 454K & 179K \\
      \# of Tokens & 175M & 53M \\
      Coverage & 97.9\% & 96.5\% \\
      \hline
    \end{tabular}
    \label{tab:lang_model}
  \end{center}
\end{table}

Each method retrieved 1,000 top documents, and the TREC evaluation
software was used to calculate non-interpolated average precision
values and plot recall-precision curves.

Table~\ref{tab:results} shows the non-interpolated average precision
values (AP) and word error rate in speech recognition, for different
retrieval methods. As with existing experiments for speech
recognition, word error rate (WER) is the ratio between the number of
word errors (i.e., deletion, insertion, and substitution) and the
total number of words. In addition, we investigated error rate
with respect to query terms (i.e., keywords used for retrieval), which
we shall call ``term error rate (TER)''.  Table~\ref{tab:results} also
shows trigram test-set perplexity (PP) and test-set out-of-vocabulary
rate (OOV).

It should noted that for all the evaluation measures in
Table~\ref{tab:results} excepting average precision, smaller values
are generally obtained with better methods.  Suggestions which can be
derived from these results are as follows.

\begin{table*}[htbp]
  \begin{center}
    \caption{Results for different retrieval methods targeting the
    NTCIR/IREX collections (AP: average
    precision, WER: word error rate, TER: term error rate,
    PP: trigram test-set perplexity, 
    OOV: test-set Out-of-Vocabulary rate).}
    \medskip
    \leavevmode
    \footnotesize
    \tabcolsep=5pt
    \begin{tabular}{l|ccccc|ccccc} \hline
      & \multicolumn{5}{|c|}{NTCIR} & \multicolumn{5}{|c}{IREX} \\
      \cline{2-11}
      \multicolumn{1}{c|}{Language Model}
      & AP & WER & TER & PP & OOV
      & AP & WER & TER & PP & OOV \\ \hline
      Text & 0.337 & --- & --- & --- & --- 
           & 0.367 & --- & --- & --- & --- \\
      NTCIR & 0.261 & 18.6\% & 23.6\% & 60  & 4.2\%
            & 0.166 & 31.1\% & 41.0\% & 138 & 6.1\% \\
      IREX  & 0.111 & 41.4\% & 54.6\% & 195 & 9.4\%  
            & 0.334 & 19.5\% & 22.9\% & 108 & 1.4\% \\
      \hline
    \end{tabular}
    \label{tab:results}
  \end{center}
\end{table*}

First, by comparing results of different language models, one can see
that the performance was significantly improved with a language model
produced from the target collection, which was observable irrespective
of the domain. Thus, producing language models based on target
collections was quite effective for speech-driven text retrieval.

Second, while in the case of the NTCIR collection, the average
precision for speech-driven retrieval was approximately 77\% of that
obtained with text-to-text retrieval, in the case of the IREX
collection, the average precision for speech-driven retrieval was
quite comparable that obtained with text-to-text retrieval.

Third, TER was generally higher than WER irrespective of the speaker.
In other words, speech recognition for content words was more
difficult than functional words, which were not contained in query
terms.

\begin{figure}[htbp]
  \begin{center}
  \leavevmode \psfig{file=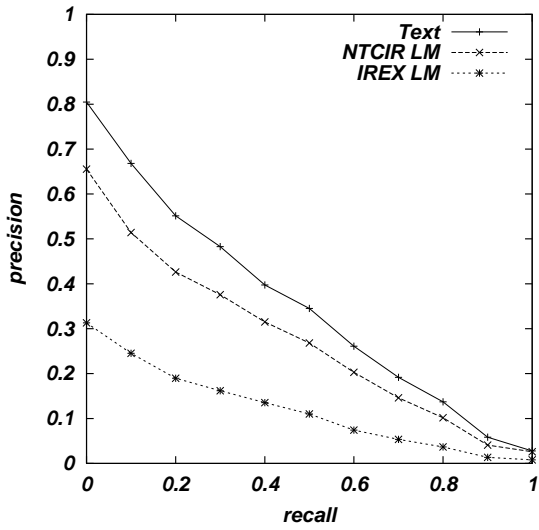,height=2.8in}
  \end{center}
  \caption{Recall-precision curves for different methods targeting the
  NTCIR collection.}
  \label{fig:ntcir}
\end{figure}

\begin{figure}[htbp]
  \begin{center}
  \leavevmode \psfig{file=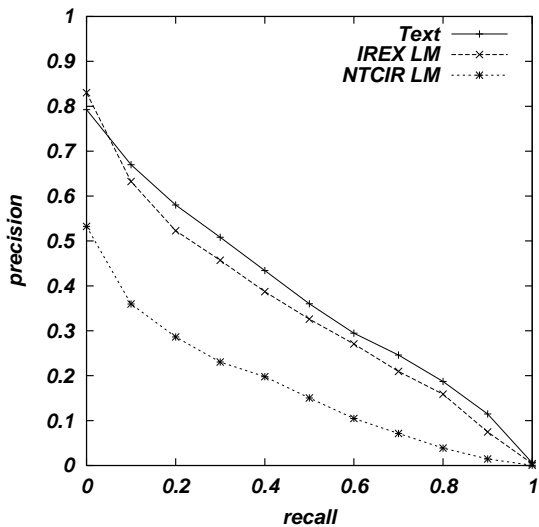,height=2.8in}
  \end{center}
  \caption{Recall-precision curves for different methods targeting the
  IREX collection.}
  \label{fig:irex}
\end{figure}

Finally, we investigated the trade-off between recall and precision.
Figures~\ref{fig:ntcir} and \ref{fig:irex} show recall-precision
curves of different retrieval methods, for the NTCIR and IREX
collections, respectively. In these figures, the relative superiority
for precision values due to different language models in
Table~\ref{tab:results} was also observable, regardless of the recall.


\section{Conclusion}
\label{sec:conclusion}

Aiming at speech-driven text retrieval with a high accuracy, we
proposed a method to integrate speech recognition and text retrieval
methods, in which target text collections are used to produce
statistical language models for speech recognition.  We also showed
the effectiveness of our method by way of experiments, where dictated
information needs in the NTCIR/IREX collections were used as queries
to retrieve documents in different domains.

\section*{Acknowledgments}

The authors would like to thank the National Institute of Informatics
for their support with the NTCIR collection and the IREX committee for
their support with the IREX collection.

\bibliographystyle{IEEEbib}

\end{document}